\pdfoutput=1
\documentclass[pdflatex,sn-mathphys-num,Numbered]{sn-jnl}

\usepackage{graphicx}%
\usepackage{amsmath,amssymb,amsfonts}%
\usepackage{booktabs}%
\usepackage{array}%
\usepackage{xcolor}%
\usepackage{pifont}%
\usepackage{url}%

\newif\ifanon
\anonfalse

\newcommand{\pp}{\,pp}
\newcommand{\MonValN}{475}
\newcommand{\MonValCtrlCorrect}{144}
\newcommand{\MonValIntCorrect}{41}
\newcommand{\MonValAbst}{186}
\newcommand{\MonValCaught}{100}
\newcommand{\MonValPrec}{0.54}
\newcommand{\MonValBase}{0.70}
\newcommand{\MonValExp}{130}

\begin{document}

\title[Structured validation, error propagation and fabricated tool use]{Measuring What Tool-Using LLM Agents Actually Do: Structured Validation, Causal Error Propagation, and Fabricated Tool Use}

\ifanon
\else
\author*[1]{\fnm{Bhaskar} \sur{Gurram}}\email{gurrambhaskar.ai@gmail.com}
\affil*[1]{\orgname{Independent Researcher}, \orgaddress{\city{Ashburn}, \state{VA}, \country{USA}}}
\fi

\abstract{Tool-using agents built on large language models (LLMs) are usually scored by comparing the final answer with a reference text, using a string heuristic or an LLM judge. This is unreliable when the references describe the task rather than its answer. We replace it with a deterministic validator that checks the tools called, the executed parameters, the resulting state, and whether the values in the answer are supported by the correct tool observations. We re-scored 33,583 traces from 13 agents on simulated calendar, weather, medical and retail tools. The simulator reproduced all 10,396 logged observations, and the gold observations supported the validator in 16 of 18 disagreements with a three-annotator majority (the annotators in none). The instrument changes conclusions: on the same 306 traces, Gemini-2.0-Flash scores 63.4\% under the substring heuristic and 5.6\% under the validator. In a paired design (same task and model, with and without one schema-valid wrong argument injected into the first tool call), the error rate rises from 52.4\% to 74.5\% (+22.0 percentage points, 95\% CI 18.4--25.5; 1,085 pairs, 7 models), and by 44.7 points (38.4--51.0) where the corrupted call executed without errors. An LLM-judge ensemble estimates +2.5 points on the same pairs. Agents also report tool results without executing any call, in 0--42.5\% of traces depending on the model (rule-based lower bounds), largely because the harness accepts an action and a final answer in one completion. A rule-based runtime monitor costs about 7\,\textmu s per trace but reduces wrong answers mainly by abstaining.}


\maketitle

\section{Introduction}\label{sec:intro}

An agent that answers questions by calling tools is an information system: it turns a request into structured queries, executes them against a data source, and reports what came back. Tool-using agents built on large language models (LLMs) follow this pattern \cite{yao2023react,schick2023toolformer}, and their reliability is normally measured by comparing the final answer with a reference, either with a string-matching heuristic or with another LLM acting as a judge \cite{zheng2023judging,gu2024surveyjudge}. Such scores are used to rank models and to decide whether a safeguard helps.

This article asks what such scores measure when the reference is not a gold value. In the benchmark studied here, as in many synthetic tool-use suites, the stored references describe the task (``Weather analysis for Mumbai'', ``Event deleted'') rather than state the correct answer. A judge that compares an answer with a task description cannot tell whether the reported temperature is the one the tool returned, whether the agent called the tool at all, or whether a scheduled appointment exists. We show that this problem is large enough to change rankings, to hide most of a causal effect, and to make a runtime safeguard look useful when it is not.

We make the evaluation checkable instead. The tools are deterministic simulators, so the correct observation for every verifiable task can be recomputed. A structured validator compares the logged trace with a gold specification derived from the query: the tools called, the parameters actually executed, the resulting state for write tasks, and whether every typed value in the answer is grounded in the gold observation. No LLM and no human label is part of the instrument.

With this instrument we study three questions on the same traces:
\begin{itemize}
\item \textbf{RQ1 (instrument validity).} How do reference-based scores (a substring heuristic and LLM judges) compare with structured validation, and does the choice change conclusions?
\item \textbf{RQ2 (causal propagation).} When one argument of the first tool call is replaced by a wrong but schema-valid value, how much does the error rate of the final answer increase, measured on paired runs?
\item \textbf{RQ3 (fabricated tool use).} How often do agents report tool-derived results without executing any tool, and why?
\end{itemize}
A secondary question concerns a rule-based runtime monitor that abstains on suspicious traces: is it better than abstaining at random?

The contributions, in order of the strength of the evidence, are:
\begin{enumerate}
\item \emph{Evaluation validity.} A deterministic, gold-grounded validator for tool-agent traces (Section~\ref{sec:validator}), with replay verification, unit tests, adjudicated agreement with human labels and a sensitivity analysis. We show that reference-based scoring misranks models (Kendall's $\tau=0.14$ between heuristic and validator rankings over seven models) and that LLM-judge agreement with humans ranges from $\kappa=0.21$ to $0.72$ across 17 prompt and judge configurations. An audit of the task files found that the stored gold call sequence contradicts the query in 360 of 945 verifiable tasks; the validator derives the gold from the query.
\item \emph{Causal propagation.} A paired estimate of the effect of a single wrong argument: +22.0\pp{} [18.4, 25.5] over 1,085 pairs and +44.7\pp{} [38.4, 51.0] where the corrupted call executed cleanly (Section~\ref{sec:rq2}). The same pairs scored by an LLM-judge ensemble give +2.5\pp{} [$-0.2$, 5.6] and by the substring heuristic $-3.2$\pp{}. Two models show no effect for identifiable reasons: Gemini-2.0-Flash rarely executes tools and GPT-4o-mini mostly emits calls that fail validation.
\item \emph{Fabricated tool use.} Rates for all 13 models, the mechanism in the harness, and a cross-check against the validator's outcome codes (Section~\ref{sec:rq3}).
\item \emph{Runtime monitor.} A negative but informative result: a monitor with no LLM calls lowers the share of wrong answers mainly through the volume of abstentions, and is no better than random abstention under the validator (Section~\ref{sec:monitor}).
\end{enumerate}
All results are conditional on the harness and the simulated tools described in Section~\ref{sec:benchmark}; Section~\ref{sec:limitations} lists what does not generalise.

\ifanon
\else
Versions 1 and 2 of this preprint used reference-based judges and a stage-wise decomposition of propagation. The re-analysis reported here replaces those measurements; Section~\ref{sec:changes} lists the claims that changed.
\fi

\section{Related work}\label{sec:related}

\paragraph{Tool-use benchmarks.} End-to-end benchmarks report task success for tool-using agents: AgentBench evaluates 27 LLMs in eight environments \cite{liu2024agentbench}, $\tau$-bench checks the final database state in tool--agent--user conversations \cite{yao2025taubench} and $\tau^2$-bench separates reasoning from coordination errors \cite{barres2025tau2}. BFCL scores individual function calls against abstract syntax trees \cite{patil2025bfcl}, and ToolSandbox evaluates stateful conversational tool use with milestones \cite{lu2025toolsandbox}. Benchmarks on real Model Context Protocol servers use LLM judges or rubric scoring \cite{wang2025mcpbench,mo2026livemcpbench}. Our validator is closest to state-based checking in $\tau$-bench, extended with parameter checks and a faithfulness check of the reported values.

\paragraph{Evaluator validity.} LLM judges are widely used \cite{zheng2023judging,gu2024surveyjudge}, and their alignment with humans varies with the judge and the task \cite{thakur2025judging}. For agents, ToolEmu checked an LM-based safety evaluator against human evaluation \cite{ruan2024toolemu}, Agent-as-a-Judge compared agentic judges with humans on code tasks \cite{zhuge2025agentasajudge}, and AgentRewardBench and Online-Mind2Web did so for web agents \cite{lu2025agentrewardbench,xue2025illusion}. A concurrent audit of tool-calling benchmarks reports substantial evaluator--human disagreement and substring-based failures \cite{bhat2026benchmarkingbenchmarks}. We add an instrument that does not depend on references at all and show how the choice of instrument changes a causal estimate.

\paragraph{Error injection and propagation.} Tools Fail injects silent errors into tool outputs \cite{sun2024toolsfail}; RoTBench perturbs tool documentation and names \cite{ye2024rotbench}; CRITICTOOL measures self-critique after tool errors \cite{huang2025critictool}. Failure attribution assigns naturally occurring failures to agents or steps \cite{zhang2025whowhen,zhu2025whereagentsfail,liu2026agenthallu}, and MAS-FIRE injects faults into multi-agent systems \cite{jia2026masfire}. Concurrent work perturbs tool-calling stages \cite{zheng2026toolrobustbench}, localises errors in parallel tool use \cite{guan2026pararecover} and questions the evidential validity of fault-injection tests \cite{he2026sscbench}. We inject a schema-valid wrong argument and estimate its effect against a paired clean run of the same task.

\paragraph{Tool hallucination and fabrication.} ToolBeHonest tests unsolvable or incomplete toolsets \cite{zhang2024toolbehonest}, Relign trains models to defer \cite{xu2025relign}, and reasoning-enhanced models have been found to hallucinate tools more often \cite{yin2026reasoningtrap}; a survey places tool-use hallucinations in a broader taxonomy \cite{lin2025agenthallucsurvey}, and MIRAGE-Bench evaluates unfaithful agent actions \cite{zhang2025miragebench}. Closest to our fabrication measurement are signed tool-execution receipts that detect fabricated tool references \cite{basu2026toolreceipts} and two later studies of fabricated outputs and assertions after tool failure \cite{soni2026toolfailbench,sethi2026fabrication}. We measure fabrication without any tool call, relate it to a specific harness design, and cross-check it with a trace-based outcome code.

\paragraph{Runtime safeguards and abstention.} Runtime enforcement and guard systems include AgentSpec \cite{wang2026agentspec}, ShieldAgent \cite{chen2025shieldagent}, GuardAgent \cite{xiang2025guardagent} and capability-based data-flow control \cite{debenedetti2026camel}; AgentDojo evaluates defences with and without attacks on the same tasks \cite{debenedetti2024agentdojo}. Abstention in LLMs has been surveyed \cite{wen2025knowyourlimits}. Following that literature, we evaluate our monitor as selective prediction, against random abstention at the same coverage.

Each of these elements (injection, evaluator validation, fabrication measurement, runtime safeguards, paired controls) is covered by some prior or concurrent work. What this study adds is their combination on shared traces and, more importantly, a measurement of how much the conclusions depend on the correctness instrument.

\section{Benchmark and harness}\label{sec:benchmark}

\paragraph{Tasks and tools.} The benchmark contains 2,000 core tasks (weather 674, calendar 667, medical 559, knowledge questions from HaluEval 100) and 300 retail tasks. Each domain exposes three tools implemented as Python functions with parameter validation: \texttt{get\_weather}, \texttt{get\_forecast}, \texttt{get\_historical}; \texttt{create\_event}, \texttt{get\_events}, \texttt{delete\_event}; \texttt{get\_patient\_record}, \texttt{check\_drug\_interaction}, \texttt{schedule\_appointment}; and \texttt{search\_products}, \texttt{get\_product}, \texttt{place\_order}. Stateless tools draw values from a random generator seeded by the call parameters, so they are deterministic functions of their arguments. The calendar uses a store shared by concurrently running sessions and is therefore not deterministic (Section~\ref{sec:validator}). Patients and drugs are synthetic. The baseline and injection experiments use a 450-task test sample of the core tasks; 1,091 distinct tasks occur in the analysed traces.

\paragraph{Models.} Thirteen agents were evaluated: nine proprietary models (GPT-4o, GPT-4o-mini, GPT-4.1-mini, GPT-4.1-nano, GPT-3.5-Turbo, o3-mini, Gemini-2.0-Flash, Gemini-2.5-Flash, DeepSeek-V3) and four open-weight models served with vLLM (Qwen2.5-3B, Qwen2.5-7B, Qwen2.5-14B in an AWQ-quantised build, and Hermes-3-Llama-3.1-8B), all at temperature 0. Two further Gemini models returned only API errors and are excluded.

\paragraph{Harness.} All agents run in the same prompted ReAct loop \cite{yao2023react}. The prompt lists tool names and descriptions but not their parameter schemas, and shows the call format with the placeholder \texttt{\{"param": "value"\}}. The parser gives priority to a \texttt{Final Answer:} line and no stop sequence is set, so a model that writes an action and a final answer in the same completion ends the episode without the action being executed. Both properties matter later: missing schemas produce parameter-name errors (e.g.\ \texttt{duration} for \texttt{duration\_minutes}, or the literal key \texttt{param}), and the parser rule produces answers with no tool execution. We report all results as conditional on this harness.

\paragraph{Error injection.} In the main injection condition (``semantic-wrong''), the first tool call of an episode is intercepted after the model has generated it and before it is executed, and one semantically critical argument is replaced by a wrong but schema-valid value: a different city (London $\to$ Manchester), another patient identifier, a date shifted by 180 days, a related drug, or a non-existent event identifier. The agent receives the observation of the corrupted call and continues. Because the substitution happens after generation, the agent cannot reject the corrupted argument; it can only notice the wrong observation and act on it. The logs record both the emitted and the executed parameters. Three schema-invalid injection types were also run and are reported in the released results.

\paragraph{Runs and traces.} Each model was run once on the 450 baseline tasks (o3-mini on 50) and once on 200 semantic-wrong tasks (GPT-4o, Gemini-2.5-Flash and o3-mini on 50); GPT-4o and Gemini-2.5-Flash have no baseline runs. Together with the monitor experiment, repeated runs and the remaining injection types, 33,583 traces were re-scored for this article: 20,910 with full step logs and 12,673 for which only the final answer and step count survive (all open-weight runs).

\paragraph{References and human labels.} The stored references are not gold values. Of the 356 non-empty references in the baseline test sample, 265 are templated task descriptions and 91 repeat the query (``Completed: \ldots''); none contains a value that is not already in the query, and the 94 knowledge tasks have empty references. Three annotators labelled 100 semantic-wrong traces from the proprietary models (10--12 per model, three domains) as correct or incorrect against these references; agreement is Fleiss' $\kappa=0.877$ [0.79, 0.95] \cite{fleiss1971kappa}, pairwise Cohen's $\kappa$ 0.835--0.939 \cite{cohen1960kappa}, with 91 of 100 traces unanimous. The majority label serves as the human reference. High agreement shows that the labels are consistent, not that they are correct: the annotators saw the same descriptive references as the judges.

\section{Structured validator}\label{sec:validator}

\paragraph{Gold specification.} For each task the validator needs the set of required calls with their scored parameters, any ordering or conditions, and the values the answer must report. The task files contain a gold call sequence, but an audit showed that it often contradicts the query the agent saw: the task generators drew slot values with a second random call instead of reusing the values rendered into the query, ignored conditions such as ``if it is in stock'', or stored a fixed placeholder call. Of the 945 verifiable tasks, 360 (38\%) have a contradicting task-file gold, most often in the product identifier (96 tasks), the number of forecast days (91), the tool multiset (58) and the appointment date (57). The validator therefore builds the gold from the query, with one parser per generator template (38 verifiable templates) and task-file values only where the query is silent. Derived requirements are computed from the deterministic simulator, never from the task file: for example, which of two products is cheaper, whether an item is in stock, or which medications a patient takes. Tasks whose required slots cannot be determined from the query are excluded as not verifiable: the 94 knowledge tasks, 25 tasks with relative dates (``tomorrow''), 15 underspecified calendar tasks, 7 appointment tasks without doctor or date, and 5 questions about side effects that no tool returns.

\paragraph{Components.} Table~\ref{tab:validator} defines the four components. $P$ uses the parameters actually executed, so an injected value that is never corrected fails $P$. $A$ extracts typed values from the answer with regular expressions and requires that every required value is present and every typed mention is grounded in the gold observation, in values from the query, or in simple aggregates of list observations (mean, minimum, maximum, range) and unit conversions. Observations of injected or wrong-entity calls are never part of the grounding set, so a value copied from a corrupted observation fails $A$. For calendar tasks, whose store is shared across sessions, the gold observation is the one the agent received from the call that matches the gold parameters. Outcome codes give the first failing condition and are used to describe errors.

\begin{table}[!tbp]
\centering
\caption{Components of the structured validator (version 1.0). Every component is computed from the query-grounded gold specification, deterministic simulator replay and the logged trace; no LLM or human judgement is involved.}
\label{tab:validator}
\small
\begin{tabular}{@{}lp{0.72\linewidth}@{}}
\toprule
Component & Definition \\
\midrule
$T$ (tools) & Every required gold call is matched one-to-one by an executed (schema-valid) call of the same tool, and per-template order constraints hold (e.g.\ \texttt{get\_events} before a conditional \texttt{create\_event}). \\
$P$ (parameters) & Every matched call agrees with its gold call on every scored slot, using the parameters actually executed (after injection). Normalisation: case, whitespace and accents; ISO dates; 24-hour times; unordered sets for attendees and drug pairs; enum spelling. \\
$S$ (state) & Write tasks only. The multiset of successful state-changing writes equals the gold writes. Conditional writes (slot free, in stock, cheaper item) are resolved from the simulator or the received observation. \\
$A$ (answer) & Typed values in the final answer (temperatures, humidity, sky condition, identifiers, prices, dates, times, drugs, conditions, severity, product and event names) are compared with the gold observation: every required value present and every typed mention grounded (derived aggregates and unit conversions allowed; $\pm1^{\circ}$). Claims about writes must match $S$. \\
CORRECT & $T\wedge P\wedge S\wedge A$; $S$ is vacuous for read-only tasks. \\
\midrule
Outcome code & First matching rule: \texttt{infra\_error} (API error text, excluded) $\to$ \texttt{correct} $\to$ \texttt{fabricated} (asserts values or a successful write although no call was executed, or claims a write that never succeeded) $\to$ \texttt{schema\_error} (every call failed validation, nothing asserted) $\to$ \texttt{refused} $\to$ \texttt{incomplete} ($\neg T$) $\to$ \texttt{wrong\_params} ($\neg P$) $\to$ \texttt{wrong\_state} ($\neg S$) $\to$ \texttt{wrong\_answer\_values} ($\neg A$). Tasks without a recoverable gold are \texttt{not\_verifiable} and excluded. \\
\bottomrule
\end{tabular}
\end{table}

\paragraph{Verification.} Three checks support the instrument. First, every successful stateless call stored in any trace was replayed through the simulator: 10,396 of 10,396 observations were reproduced exactly (the run-time dates in forecasts excepted). Second, 24 unit tests on hand-constructed traces cover every outcome code, the injected-value path, conditional writes, unit conversions and the task-file contradiction. Third, on the 85 verifiable traces of the human-labelled sample the validator agrees with the three-annotator majority at $\kappa=0.53$ [0.32, 0.70] (Table~\ref{tab:human}); human--human agreement on the same traces is 0.85--0.95. Moderate agreement is expected, because the annotators judged against task descriptions. We therefore adjudicated all 18 disagreements against the replayed gold observations: the gold supported the validator in 16 cases, the annotators in none, and two were ambiguous (an hourly forecast delivered as a daily value; an event titled ``All Hands Meeting'' instead of ``All Hands''). Eight of the 18 are traces in which the agent executed no tool but described a plausible result; four are correct refusals to create an event in an occupied slot or to cancel a non-existent event, which the annotators marked wrong because the reference assumed the action happened; one copies the injected wrong-city value.

\begin{table}[!tbp]
\centering
\caption{Agreement with human labels on the 100 annotated P2 traces. Humans and LLM judges saw only the descriptive reference; the validator uses the gold observations. Cohen's $\kappa$ against the three-annotator majority with 95\% bootstrap CIs ($B=10{,}000$). The validator is compared on the 85 verifiable traces; the 18 disagreements were adjudicated against the replayed gold observations.}
\label{tab:human}
\small
\begin{tabular}{@{}llc@{}}
\toprule
Instrument & $n$ & $\kappa$ vs majority [95\% CI] \\
\midrule
Three annotators, Fleiss' $\kappa$ & 100 & 0.877 [0.79, 0.95] \\
Substring heuristic & 100 & $-0.011$ [$-0.19$, 0.17] \\
GPT-4o judge & 100 & 0.274 [0.09, 0.46] \\
Gemini-2.5-Flash judge & 100 & 0.372 [0.22, 0.54] \\
GPT-4o-mini judge & 100 & 0.536 [0.36, 0.69] \\
Three-LLM ensemble (majority) & 100 & 0.413 [0.24, 0.58] \\
LLM judges, 17 prompt/judge configurations & 100 & 0.21 to 0.72 (range) \\
Structured validator & 85 & 0.53 [0.32, 0.70] \\
\midrule
\multicolumn{3}{@{}l}{Validator/majority confusion: YES/YES 19, YES/NO 5, NO/YES 13, NO/NO 48} \\
\multicolumn{3}{@{}l}{Adjudication of the 18 disagreements: validator supported 16, humans 0, ambiguous 2} \\
\bottomrule
\end{tabular}
\end{table}

\paragraph{Sensitivity.} We switched each of 20 normalisation and rubric choices off or varied it, one at a time. No switch moved pooled baseline accuracy (46.8\%) by more than 2.7 points or changed more than 2.0\% of labels, and the per-model ranking was unchanged (Kendall's $\tau=1.00$) under every switch. The largest effects came from dropping time normalisation (44.1\%), dropping derived aggregates (45.0\%) and requiring only half of the required values (49.3\%). Replacing the query-grounded gold with the raw task-file sequence had a much larger effect (27.5\%, 16.4\% of labels changed, $\tau=0.90$).

\paragraph{Answer-only mode.} The open-weight runs kept only final answers. For them the validator computes $A$ alone, on 170 read tasks with deterministic tools. This mode is more lenient than the full check: on 1,040 proprietary traces scored both ways it accepts 49.4\% against 41.5\% for CORRECT. Open-weight scores are therefore not comparable with proprietary ones and are not used in the causal analysis.

\section{Experimental design}\label{sec:design}

\paragraph{Instruments.} The validator is the primary instrument. The substring heuristic (an answer is correct if the first 20 characters of the reference appear in the first 200 characters of the answer, or one of the first three reference words longer than three characters appears) and the three-LLM ensemble (majority of GPT-4o, Gemini-2.5-Flash and GPT-4o-mini, all prompted to compare the answer with the reference) are the instruments of the earlier analysis; we apply them to the same traces as sensitivity analyses. The GPT-4o-mini judge alone is also reported because it agreed best with the annotators among the original judges. The LLM judges are therefore objects of study here, not part of the measurement pipeline.

\paragraph{Paired causal design.} Each semantic-wrong trace is paired with the baseline trace of the same task and model. Both runs use the same prompt, loop and temperature; only the injection differs. A pair is included if the task is verifiable and neither run ended in an API error. This yields 1,085 pairs over seven models (175 pairs each, 35 for o3-mini). The estimand is the risk difference $\mathrm{RD}=P(\text{wrong}\mid\text{injected})-P(\text{wrong}\mid\text{clean})$. We report exact McNemar tests on the discordant pairs \cite{mcnemar1947} and 95\% percentile confidence intervals from a bootstrap that resamples tasks ($B=2000$), because the same task appears for several models. We report two subsets: pairs in which the injected run executed at least one call and no call failed validation (``executed cleanly''), and pairs in which neither run executed any tool. The first subset is defined on the injected arm and is descriptive rather than causal. The second is a negative control for the LLM judges; under the validator it is wrong in both arms by construction, so it carries no information there.

\paragraph{Fabrication.} A rule-based classifier marks a trace as fabricated if no tool output was obtained (verified against the step logs) and the final answer asserts a tool-derived action or result, excluding empty answers, pure placeholders, statements of intent and refusals. The rule was fixed before validation and then compared with two LLM judges (GPT-4.1-mini and GPT-4o) on all 734 semantic-wrong traces without a tool call. These comparisons are provisional until human labels are available. The validator's \texttt{fabricated} outcome code provides an independent, trace-based count on verifiable tasks.

\paragraph{Real-tool pilot.} To bound the effect of simulated execution on fabrication, four OpenAI models ran the same 80 tasks once with simulated tools and once with real tools (DuckDuckGo search, a sandboxed Python interpreter, a calculator and whitelisted file reads). The pilot uses a separate task suite (question answering, code, web navigation, data analysis and reasoning). Equivalence was tested with two one-sided tests at a margin of $\pm10$\pp{} fixed in advance \cite{schuirmann1987tost}, using Newcombe's paired intervals \cite{newcombe1998paired}.

\paragraph{Other statistics.} Proportions carry Wilson intervals \cite{wilson1927}; agreement is Cohen's $\kappa$ \cite{cohen1960kappa} or Fleiss' $\kappa$ \cite{fleiss1971kappa} with bootstrap intervals ($B=10{,}000$), interpreted following \cite{landis1977kappa}. All analysis scripts open the trace databases read-only and are released with the data.

\paragraph{Why the stage decomposition was dropped.}\label{sec:hop}
The earlier analysis decomposed propagation into stages: whether the injection was applied, whether it changed execution (an ``error propagation stage'' of at least 2), and whether the answer was wrong, and reported a hop rate $r_{23}$ between the last two. Reading the scoring code showed that this decomposition does not measure propagation. The stage score is the product of two indicators: the first action names a valid tool (only then does the injector run), and either an observation contains a validation or tool error or the heuristic judges the answer wrong. The second stage therefore requires heuristic failure, so $r_{23}=1$ by construction under the heuristic, and it selects traces whose own malformed calls failed validation, in which the injected value mostly never executed (the substituted value appears in 1.9\% of these answers, against 22.1\% when the corrupted call executed without error). Across models, the apparent first-stage ``rejection'' rate mostly measured whether the agent emitted a tool call at all ($\rho=-0.97$ with the tool-call admission rate), not whether it rejected corrupted input. We replaced the decomposition by the paired design above and withdrew the hop rate and the reported decoupling of rejection and recovery.

\section{Results}\label{sec:results}

\subsection{RQ1: instrument validity}\label{sec:rq1}

\begin{figure}[!tbp]
\centering
\includegraphics[width=\linewidth]{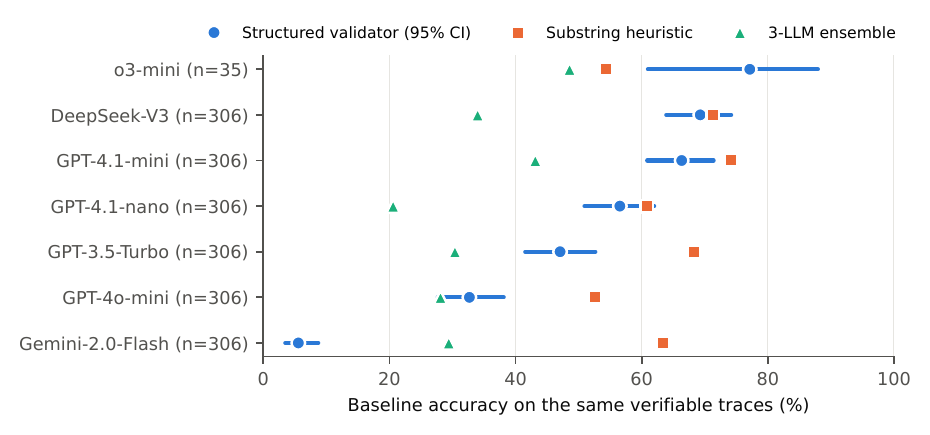}
\caption{Baseline accuracy of seven models under three instruments applied to the same verifiable traces. Horizontal lines are Wilson 95\% intervals for the validator.}
\label{fig:accuracy}
\end{figure}

Table~\ref{tab:accuracy} and Figure~\ref{fig:accuracy} compare the three instruments on the same verifiable baseline traces. The validator spreads the seven models from 5.6\% (Gemini-2.0-Flash) to 77.1\% (o3-mini). The heuristic places all of them between 52.6\% and 74.2\% and orders them almost unrelatedly (Kendall's $\tau=0.14$); the ensemble is lower everywhere and only partly consistent with the validator ($\tau=0.52$). The largest discrepancy is Gemini-2.0-Flash: 63.4\% under the heuristic and 5.6\% [3.5, 8.7] under the validator. It executes no tool in 94\% of its verifiable baseline traces because it writes the action and the final answer in one completion, and 51.0\% of its traces are coded as fabricated. The heuristic accepts these answers because they mention the city or the event title that also appears in the reference.

\begin{table}[!tbp]
\centering
\caption{Baseline accuracy (\%) of the seven proprietary models with stored baseline traces, scored by three instruments on the same verifiable traces. Validator: structured validator with Wilson 95\% CI and component pass rates ($T$ tools, $P$ parameters, $A$ answer faithfulness; $S$ is reported in the released results). Fab.: share of traces with outcome code \texttt{fabricated}. Heuristic and ensemble: the labels used in the earlier version of this study, restricted to the same traces. Kendall's $\tau$ between the validator ranking and the heuristic ranking is 0.14; with the ensemble ranking it is 0.52.}
\label{tab:accuracy}
\footnotesize
\setlength{\tabcolsep}{3.5pt}
\begin{tabular}{@{}lrcrrrrrr@{}}
\toprule
Model & $n$ & Validator [95\% CI] & $T$ & $P$ & $A$ & Fab. & Heuristic & Ensemble \\
\midrule
o3-mini & 35 & 77.1 [61.0, 87.9] & 82.9 & 80.0 & 80.0 & 5.7 & 54.3 & 48.6 \\
DeepSeek-V3 & 306 & 69.3 [63.9, 74.2] & 77.8 & 73.9 & 77.5 & 8.2 & 71.2 & 34.0 \\
GPT-4.1-mini & 306 & 66.3 [60.9, 71.4] & 73.5 & 72.5 & 68.6 & 12.7 & 74.2 & 43.1 \\
GPT-4.1-nano & 306 & 56.5 [50.9, 62.0] & 65.0 & 61.1 & 62.1 & 0.0 & 60.8 & 20.6 \\
GPT-3.5-Turbo & 306 & 47.1 [41.5, 52.7] & 51.3 & 47.7 & 52.9 & 11.8 & 68.3 & 30.4 \\
GPT-4o-mini & 306 & 32.7 [27.7, 38.1] & 44.4 & 43.1 & 35.3 & 12.1 & 52.6 & 28.1 \\
Gemini-2.0-Flash & 306 & 5.6 [3.5, 8.7] & 5.9 & 5.9 & 7.5 & 51.0 & 63.4 & 29.4 \\
\bottomrule
\end{tabular}
\end{table}

Against the human majority, the heuristic is at chance ($\kappa=-0.011$) and the original LLM judges reach $\kappa=0.27$--$0.54$ (Table~\ref{tab:human}). The single GPT-4o-mini judge agrees better than the ensemble (paired difference in $\kappa$ +0.12 [0.004, 0.25]). Agreement depends strongly on wording: re-running four OpenAI judges with four prompts gave $\kappa$ between 0.21 and 0.72 across 17 configurations (mean 0.49, SD 0.15), with a range of 0.23--0.39 across prompts for a single judge, and the best prompt was written after inspecting this sample. All judges were conservative relative to the annotators. We found no vendor-level self-preference; GPT-4o was somewhat more lenient on its own traces (+0.12 [0.01, 0.24], $n=50$), which is consistent with self-preference but underpowered. String metrics computed against the references (exact match, token overlap, numeric tolerance) reach at most $\kappa=0.32$, which is expected when no reference contains the answer.

Two observations summarise RQ1. Reference-based instruments, human or automatic, judge whether an answer is consistent with a description of the task; they cannot see whether the reported values came from the right call. And because the judges disagree with each other and depend on prompt wording, they cannot serve as a fixed yardstick for comparing conditions, which is what the next analysis requires.

\subsection{RQ2: causal effect of a wrong argument}\label{sec:rq2}

\begin{figure}[!tbp]
\centering
\includegraphics[width=\linewidth]{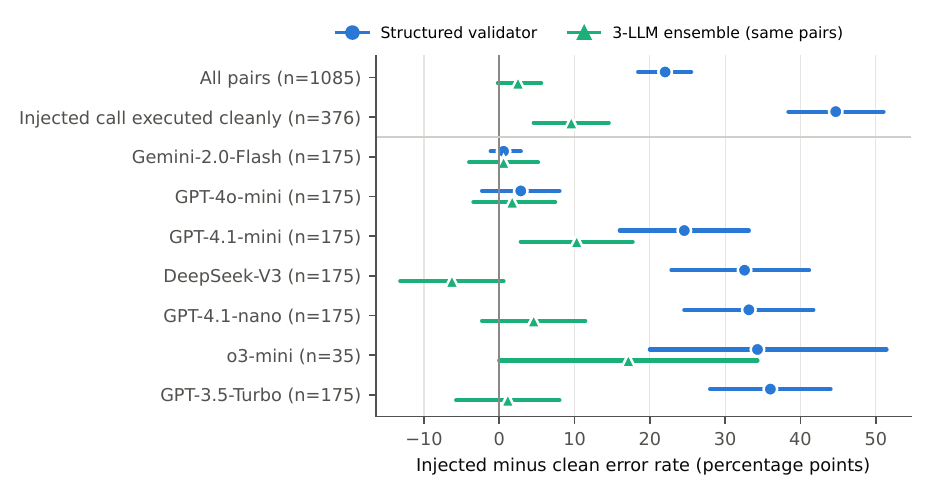}
\caption{Paired effect of one schema-valid wrong argument, by stratum and model, under the validator and under the three-LLM ensemble on the same pairs. Intervals are 95\% task-cluster bootstrap CIs.}
\label{fig:forest}
\end{figure}

Table~\ref{tab:paired} and Figure~\ref{fig:forest} give the paired results. Under the validator, injection raises the error rate from 52.4\% to 74.5\% (RD +22.0\pp{} [18.4, 25.5]; 286 pairs wrong only when injected against 47 wrong only when clean; McNemar $p=6\times10^{-43}$). Of the 516 pairs answered correctly without injection, 286 (55.4\% [51.1, 59.7]) become wrong with it. Where the injected run executed at least one call and none failed validation ($n=376$), the error rate rises from 26.3\% to 71.0\% (+44.7\pp{} [38.4, 51.0]), and 186 of 277 correctly answered tasks (67.1\% [61.4, 72.4]) become wrong.

\begin{table}[!tbp]
\centering
\caption{Paired effect of injecting one schema-valid wrong argument into the first tool call. Each pair is the same task run by the same model without and with injection. Err.: error rate (\%) in the clean and injected arm; RD: risk difference in percentage points with a 95\% task-cluster bootstrap CI ($B=2000$); $b/c$: pairs wrong only when injected / wrong only when clean; $p$: exact McNemar test. The executed-cleanly stratum (the injected run executed at least one call and no call failed validation) is defined on the injected arm and is therefore descriptive. Under the validator the no-tool stratum is wrong in both arms by construction and is not an informative placebo; the LLM-judge rows provide that check.}
\label{tab:paired}
\footnotesize
\setlength{\tabcolsep}{3pt}
\begin{tabular}{@{}lrrrlrr@{}}
\toprule
Instrument / stratum & $n$ & Err.\ clean & Err.\ inj. & RD [95\% CI] & $b/c$ & $p$ \\
\midrule
\multicolumn{7}{@{}l}{\emph{Structured validator}} \\
\quad All pairs & 1085 & 52.4 & 74.5 & +22.0 [+18.4, +25.5] & 286/47 & $6.3\times10^{-43}$ \\
\quad Injected call executed cleanly & 376 & 26.3 & 71.0 & +44.7 [+38.4, +51.0] & 186/18 & $2.3\times10^{-36}$ \\
\quad No tool executed in either arm & 351 & 100.0 & 100.0 & +0.0 [+0.0, +0.0] & 0/0 & -- \\
\multicolumn{7}{@{}l}{\emph{3-LLM ensemble}} \\
\quad All pairs & 1085 & 67.2 & 69.7 & +2.5 [-0.2, +5.6] & 118/91 & 0.07 \\
\quad Injected call executed cleanly & 376 & 80.1 & 89.6 & +9.6 [+4.6, +14.5] & 55/19 & $3.4\times10^{-5}$ \\
\quad No tool executed in either arm & 351 & 60.7 & 57.8 & -2.8 [-6.6, +0.6] & 16/26 & 0.16 \\
\multicolumn{7}{@{}l}{\emph{GPT-4o-mini judge}} \\
\quad All pairs & 1085 & 51.4 & 56.0 & +4.6 [+1.2, +8.1] & 156/106 & 0.002 \\
\quad Injected call executed cleanly & 376 & 59.6 & 73.1 & +13.6 [+7.6, +19.8] & 85/34 & $3.3\times10^{-6}$ \\
\quad No tool executed in either arm & 351 & 47.9 & 43.9 & -4.0 [-7.8, +0.0] & 15/29 & 0.05 \\
\multicolumn{7}{@{}l}{\emph{Substring heuristic}} \\
\quad All pairs & 1085 & 35.0 & 31.8 & -3.2 [-6.0, -0.4] & 76/111 & 0.01 \\
\quad Injected call executed cleanly & 376 & 36.7 & 27.1 & -9.6 [-15.4, -4.0] & 26/62 & $1.6\times10^{-4}$ \\
\quad No tool executed in either arm & 351 & 31.3 & 31.3 & +0.0 [-2.9, +3.0] & 13/13 & 1.00 \\
\midrule
\multicolumn{7}{@{}l}{\emph{Structured validator, per model (all pairs)}} \\
\quad GPT-3.5-Turbo & 175 & 53.1 & 89.1 & +36.0 [+28.0, +44.0] & 67/4 & $8.7\times10^{-16}$ \\
\quad o3-mini & 35 & 22.9 & 57.1 & +34.3 [+20.0, +51.4] & 12/0 & $4.9\times10^{-4}$ \\
\quad GPT-4.1-nano & 175 & 42.3 & 75.4 & +33.1 [+24.6, +41.7] & 65/7 & $7.0\times10^{-13}$ \\
\quad DeepSeek-V3 & 175 & 29.7 & 62.3 & +32.6 [+22.9, +41.1] & 71/14 & $2.4\times10^{-10}$ \\
\quad GPT-4.1-mini & 175 & 33.7 & 58.3 & +24.6 [+16.0, +33.1] & 56/13 & $1.7\times10^{-7}$ \\
\quad GPT-4o-mini & 175 & 66.9 & 69.7 & +2.9 [-2.3, +8.0] & 13/8 & 0.38 \\
\quad Gemini-2.0-Flash & 175 & 94.9 & 95.4 & +0.6 [-1.1, +2.9] & 2/1 & 1.00 \\
\bottomrule
\end{tabular}
\end{table}

The outcome codes show what happens in these flips. Of the 286 pairs that turn from correct to wrong, 157 end as \texttt{wrong\_params}: the agent never repeated the call with the correct argument. In 152 of these the parameters the agent itself emitted were correct, so the error is the injected one, and in 135 the answer also fails the faithfulness check, i.e.\ it reports values that the gold observation does not support. The remaining flips end as incomplete (48), refused (39), fabricated (27), wrong answer values (9) or schema errors (6). Recovery, in the sense of noticing the wrong observation and repeating the call correctly, is rare in this harness.

\begin{figure}[!tbp]
\centering
\includegraphics[width=\linewidth]{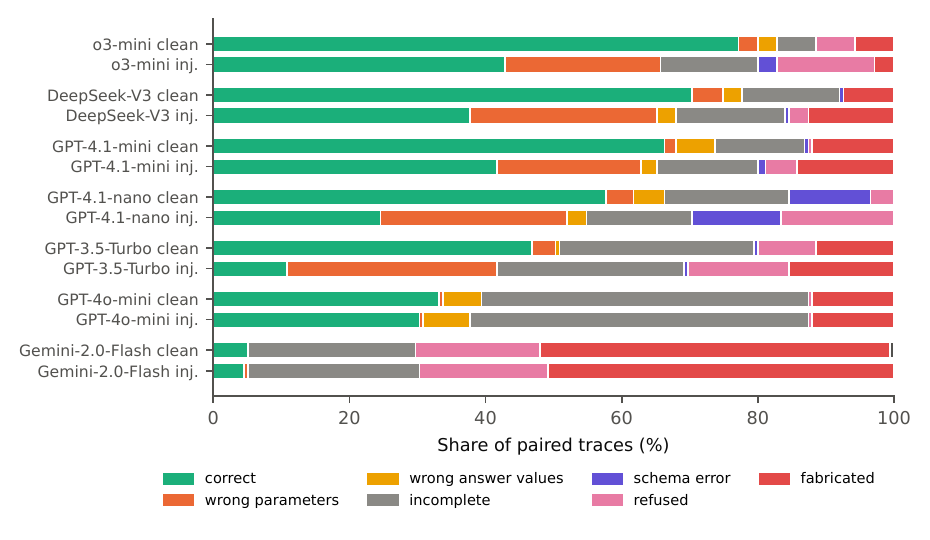}
\caption{Validator outcome codes in the clean and injected arm of the 1,085 pairs, by model. Injection mainly converts correct outcomes into wrong parameters (the corrupted call was never repeated correctly). Gemini-2.0-Flash is dominated by fabricated and incomplete outcomes in both arms.}
\label{fig:outcomes}
\end{figure}

Five of the seven models show large effects, from +24.6\pp{} (GPT-4.1-mini) to +36.0\pp{} (GPT-3.5-Turbo). Two do not, and the outcome codes explain why (Figure~\ref{fig:outcomes}). Gemini-2.0-Flash executed no tool in 165 of its 175 injected runs, so the injector rarely had a call to corrupt; its error rate is above 94\% in both arms. GPT-4o-mini emitted at least one call that failed schema validation in 79 of 175 injected runs and executed no call in 95; only one of its pairs falls in the executed-cleanly subset. A corrupted argument cannot propagate through a call that is never executed. These near-null effects are therefore findings about tool-call behaviour in this harness, not evidence of robustness.

\paragraph{Sensitivity to the instrument.} On the same 1,085 pairs the three-LLM ensemble estimates +2.5\pp{} [$-0.2$, 5.6] ($p=0.07$), and +9.6\pp{} [4.6, 14.5] in the executed-cleanly subset; the GPT-4o-mini judge estimates +4.6\pp{} [1.2, 8.1]. The substring heuristic gives the wrong sign ($-3.2$\pp{} [$-6.0$, $-0.4$]), because an answer that names the wrong city still contains words of the reference. The earlier analysis, which used the ensemble on 2,041 pairs from 11 models including the open-weight ones, reported +3.4\pp{} [1.7, 5.2] overall and +10.6\pp{} [7.3, 13.9] where the corrupted call executed silently. The LLM judges pass the negative control: in pairs where neither run executed a tool the ensemble estimates $-2.8$\pp{} [$-6.6$, 0.6], and $-0.5$\pp{} [$-2.8$, 1.7] on the larger set. They detect the direction of the effect but attenuate its size by a factor of about five to nine, mainly because they mark most answers wrong in both arms (ensemble error 67.2\% clean, 69.7\% injected) and so register few discordant pairs. The ensemble is not uniformly attenuated: for DeepSeek-V3 it estimates $-6.3$\pp{} where the validator estimates +32.6\pp{}.

\subsection{RQ3: fabricated tool use}\label{sec:rq3}

Table~\ref{tab:fabrication} reports fabrication on the semantic-wrong runs of all 13 models. The rule-based rate ranges from 0.0\% (GPT-4o, Qwen2.5-7B) to 38.5\% (Gemini-2.0-Flash) and 42.5\% (Hermes-3-Llama-3.1-8B). Across models, the rate follows the share of traces without a tool call (Spearman's $\rho=0.88$).

\begin{table}[!tbp]
\centering
\caption{Fabricated tool use on P2 semantic-wrong traces (\%, Wilson 95\% CIs). Rule: rule-based classifier (no tool output obtained and the answer asserts a tool-derived result); a lower bound, with recall 66--73\% against LLM judges. LLM-adj.: GPT-4.1-mini judge with pure placeholders counted as not fabricated (provisional). No call: fabricated among traces without a tool call (LLM-adj.). Validator: share of verifiable injected-arm traces with outcome \texttt{fabricated}; open-weight runs have no step logs.}
\label{tab:fabrication}
\footnotesize
\setlength{\tabcolsep}{3pt}
\begin{tabular}{@{}llrrllrr@{}}
\toprule
Model & Type & $n$ & Tool call & Rule [95\% CI] & LLM-adj.\ [95\% CI] & No call & Validator \\
\midrule
Gemini-2.0-Flash & prop. & 200 & 5.0 & 38.5 [32.0, 45.4] & 65.5 [58.7, 71.7] & 131/190 & 89/175 \\
GPT-3.5-Turbo & prop. & 200 & 59.5 & 18.0 [13.3, 23.9] & 23.0 [17.7, 29.3] & 46/81 & 27/175 \\
Gemini-2.5-Flash & prop. & 50 & 76.0 & 18.0 [9.8, 30.8] & 18.0 [9.8, 30.8] & 9/12 & 8/41 \\
GPT-4o-mini & prop. & 200 & 40.0 & 14.5 [10.3, 20.0] & 14.5 [10.3, 20.0] & 29/120 & 21/175 \\
GPT-4.1-mini & prop. & 200 & 85.0 & 11.0 [7.4, 16.1] & 14.0 [9.9, 19.5] & 28/30 & 25/175 \\
DeepSeek-V3 & prop. & 200 & 86.0 & 10.5 [7.0, 15.5] & 12.0 [8.2, 17.2] & 24/28 & 22/175 \\
o3-mini & prop. & 50 & 88.0 & 8.0 [3.2, 18.8] & 8.0 [3.2, 18.8] & 4/6 & 1/35 \\
GPT-4.1-nano & prop. & 200 & 99.0 & 0.5 [0.1, 2.8] & 0.5 [0.1, 2.8] & 1/2 & 0/175 \\
GPT-4o & prop. & 50 & 100.0 & 0.0 [0.0, 7.1] & 0.0 [0.0, 7.1] & -- & 0/41 \\
\midrule
Hermes-3-Llama-3.1-8B & open & 200 & 37.0 & 42.5 [35.9, 49.4] & 48.5 [41.7, 55.4] & 97/126 & -- \\
Qwen2.5-14B & open & 200 & 61.0 & 17.0 [12.4, 22.8] & 26.0 [20.4, 32.5] & 52/78 & -- \\
Qwen2.5-3B & open & 200 & 70.0 & 6.5 [3.8, 10.8] & 8.0 [5.0, 12.6] & 16/60 & -- \\
Qwen2.5-7B & open & 200 & 99.5 & 0.0 [0.0, 1.9] & 0.0 [0.0, 1.9] & 0/1 & -- \\
\bottomrule
\end{tabular}
\end{table}

\paragraph{Mechanism.} Fabrication here is a joint product of model and harness. All 469 proprietary semantic-wrong traces without a tool call received no observation, and in 462 of them the model wrote an \texttt{Action:} and a \texttt{Final Answer:} in the same completion. Because the parser gives priority to the final answer and no stop sequence is set, the episode ends before the action runs, and the answer then states the result the model expected. The same model can behave differently under another interface: Qwen2.5-14B issued tool calls in 61\% of traces in this harness and in 93.5\% under native function calling. Rates in Table~\ref{tab:fabrication} are therefore properties of models in this harness, not of models in general.

\paragraph{Classifier validity.} Against the GPT-4.1-mini judge over all 734 traces without a tool call, the rule has precision 96.4\% [93.8, 97.9] and recall 66.3\% [62.0, 70.4]; counting pure placeholders as not fabricated raises recall to 73.0\% [68.6, 76.9]. The two judges agree with each other at $\kappa=0.92$, and precision against GPT-4o is 96.7\%. The rule thus rarely flags a trace wrongly but misses about a third of the cases the judges flag; its rates are lower bounds, and the judge-based figures remain provisional until human labels are collected. The validator's trace-based \texttt{fabricated} code, computed on verifiable tasks with a different definition, gives a similar ordering (last column of Table~\ref{tab:fabrication}): 89 of 175 injected runs for Gemini-2.0-Flash, 12--15\% for GPT-3.5-Turbo, GPT-4.1-mini, DeepSeek-V3 and GPT-4o-mini, and none for GPT-4.1-nano and GPT-4o.

\paragraph{Proprietary versus open-weight.} The mean rule-based rate is 13.2\% over nine proprietary models and 16.5\% over four open-weight models (exact permutation test over models, $p=0.75$); with the adjusted judge labels the means are 17.3\% and 20.6\% ($p=0.84$). With four open-weight models the comparison has little power, and we draw no conclusion from it.

\paragraph{Fabrication and robustness scores.} Fabrication interacts with RQ2. A model that does not execute tools cannot be harmed by a corrupted argument, so injection-based robustness scores reward it. In the earlier analysis Gemini-2.0-Flash appeared to ``reject'' 95\% of injections; the validator shows that it executes almost no tools and fabricates in about half of its verifiable traces in both arms (Figure~\ref{fig:outcomes}).

\subsection{Exploratory pilot with real tools}\label{sec:pilot}

\begin{table}[!tbp]
\centering
\caption{Exploratory pilot: the same 80 tasks per model run once with simulated and once with real tools. Fabricated = no successful tool call and a confident answer. $\Delta$ = real minus simulated (pp) with Newcombe paired 95\% and 90\% CIs (pooled: task-cluster bootstrap); $b/c$: real-only / simulated-only fabrications with exact McNemar $p$. Equivalence margin $\pm10$ pp, fixed in advance (TOST).}
\label{tab:pilot}
\footnotesize
\setlength{\tabcolsep}{3.5pt}
\begin{tabular}{@{}lrrrrllll@{}}
\toprule
Model & $n$ & Sim.\ fab. & Real fab. & $\Delta$ & 95\% CI & 90\% CI & $b/c$, $p$ & Equivalent \\
\midrule
GPT-4o-mini & 80 & 10.0 & 11.2 & +1.2 & [$-5.9$, 8.6] & [$-4.6$, 7.3] & 4/3, 1.00 & yes \\
GPT-4.1-mini & 80 & 12.5 & 13.8 & +1.3 & [$-7.3$, 9.9] & [$-5.8$, 8.4] & 6/5, 1.00 & yes \\
GPT-4o & 80 & 8.8 & 6.2 & $-2.5$ & [$-9.0$, 3.3] & [$-7.7$, 2.1] & 1/3, 0.62 & yes \\
GPT-4.1 & 80 & 23.8 & 15.0 & $-8.8$ & [$-17.2$, $-0.8$] & [$-15.7$, $-2.1$] & 2/9, 0.07 & no \\
\midrule
Pooled & 320 & 13.8 & 11.6 & $-2.2$ & [$-5.6$, 1.2] & [$-5.3$, 0.6] & 13/20, 0.30 & yes \\
\bottomrule
\end{tabular}
\end{table}

Table~\ref{tab:pilot} compares fabrication with simulated and real tools on 320 paired tasks. Pooled over four models, the difference is $-2.2$\pp{} (95\% CI [$-5.6$, 1.2]; 90\% CI [$-5.3$, 0.6]), inside the $\pm10$\pp{} margin, so the two conditions are equivalent at that margin; the tool-call rate is identical in both arms (84.7\%). GPT-4.1 alone is not equivalent ($-8.8$\pp{} [$-17.2$, $-0.8$]). Of the 33 discordant pairs, 23 differ in whether the model decided to call a tool before anything was executed, which points to run-to-run variation rather than to the tool backend, and only one differs in execution. The pilot bounds backend-induced shifts in fabrication within its own pipeline; it uses a different task suite and only OpenAI models, and it does not validate the simulated tools for the injection experiments. The stronger argument for those is structural: by definition, no fabricated trace received any tool output.

\section{Runtime monitor}\label{sec:monitor}

The monitor reads only artefacts the agent already produces and makes no LLM calls. In the configuration that was run, layer 1 abstains at the second schema-validation or tool error, layer 2 logs uncertainty keywords in the reasoning text without acting on them, and layer 3 abstains if the answer contains a number that appears in no observation. It was evaluated against a task-matched control run on 600 tasks for GPT-4o-mini and Gemini-2.0-Flash (the control was run two days later) and on 300 tasks for three open-weight models with concurrent control runs. We treat it as selective prediction: abstaining always lowers the share of answered-and-wrong outputs, so the relevant comparison is with random abstention at the same coverage.

\begin{table}[!tbp]
\centering
\caption{The runtime monitor as selective prediction (task-matched control vs monitor arm). Wrong: share of tasks answered wrongly (abstentions count as not wrong), control $\to$ monitor; $\Delta$ with paired bootstrap 95\% CI. Cov.: share of tasks answered. Prec.: share of abstained tasks that the control answered wrongly (base error rate in parentheses). Random: expected wrong share under random abstention at the same rate. $p_{\mathrm{rand}}$: probability that random abstention catches at least as many errors. Labels: H substring heuristic, J GPT-4o-mini judge.}
\label{tab:monitor}
\footnotesize
\setlength{\tabcolsep}{3.5pt}
\begin{tabular}{@{}llrrlrlrr@{}}
\toprule
Model & Lab. & $n$ & Wrong \% & $\Delta$ [95\% CI] & Cov. & Prec.\ (base) & Rand. & $p_{\mathrm{rand}}$ \\
\midrule
GPT-4o-mini & H & 600 & 55.8 $\to$ 32.8 & $-23.0$ [$-27.0$,$-18.8$] & 66.5 & 0.66 (0.56) & 37.1 & $<0.001$ \\
GPT-4o-mini & J & 600 & 53.7 $\to$ 32.0 & $-21.7$ [$-26.0$, $-17.3$] & 66.5 & 0.66 (0.54) & 35.7 & $<0.001$ \\
Gemini-2.0-F. & J & 600 & 39.2 $\to$ 38.5 & $-0.7$ [$-3.0$, 1.7] & 97.2 & 0.47 (0.39) & 38.1 & 0.33 \\
Qwen2.5-7B & H & 300 & 55.3 $\to$ 31.3 & $-24.0$ [$-29.3$, $-19.0$] & 44.7 & 0.45 (0.55) & 24.7 & 1.00 \\
Qwen2.5-14B & H & 300 & 45.0 $\to$ 37.0 & $-8.0$ [$-11.3$, $-4.7$] & 73.0 & 0.30 (0.45) & 32.9 & 1.00 \\
Hermes-3-8B & H & 300 & 43.0 $\to$ 39.0 & $-4.0$ [$-7.3$, $-0.7$] & 86.7 & 0.40 (0.43) & 37.3 & 0.72 \\
\bottomrule
\end{tabular}
\end{table}

Table~\ref{tab:monitor} shows the results under the labels available for each run. For GPT-4o-mini, the monitor abstains on 33.5\% of tasks and lowers the wrong-answer share from 55.8\% to 32.8\% under the heuristic (53.7\% to 32.0\% under the GPT-4o-mini judge). Random abstention at the same rate would reach 37.1\% (35.7\%), so about 4 of the 23 points reflect discrimination and the rest reflect the number of abstentions. The discriminating part comes from layer 1 (139 abstentions, precision 0.85 against a base error rate of 0.56); layer 3 abstained 62 times at precision 0.24, below the base rate, partly because its number pattern matches list numbering and the parts of ISO dates. On the three open-weight models the monitor abstains on 13--55\% of tasks and lowers the wrong-answer share by 4--24 points, but its abstentions are no more accurate than random (precision 0.30--0.45 against base rates 0.43--0.55); on Qwen2.5-7B the answers it lets through are less accurate than the control's (29.9\% against 44.7\%). Gemini-2.0-Flash executes so few tools that the monitor rarely fires (coverage 97.2\%).

These figures rest on reference-based labels. Re-scoring the GPT-4o-mini runs with the validator on the \MonValN{} verifiable tasks gives a less favourable picture. The control is correct on \MonValCtrlCorrect{} tasks and the monitor arm on \MonValIntCorrect{}. Of the \MonValAbst{} tasks on which the monitor abstained, the control answered \MonValCaught{} wrongly (precision \MonValPrec{}), fewer than the \MonValExp{} expected from random abstention given the base error rate of \MonValBase{}. Part of this is run-to-run variation between the two arms (two identical temperature-0 runs of GPT-4o-mini disagree on about 15\% of labels), but the result does not support a claim that the monitor selects wrong answers.

\begin{figure}[!tbp]
\centering
\includegraphics[width=\linewidth]{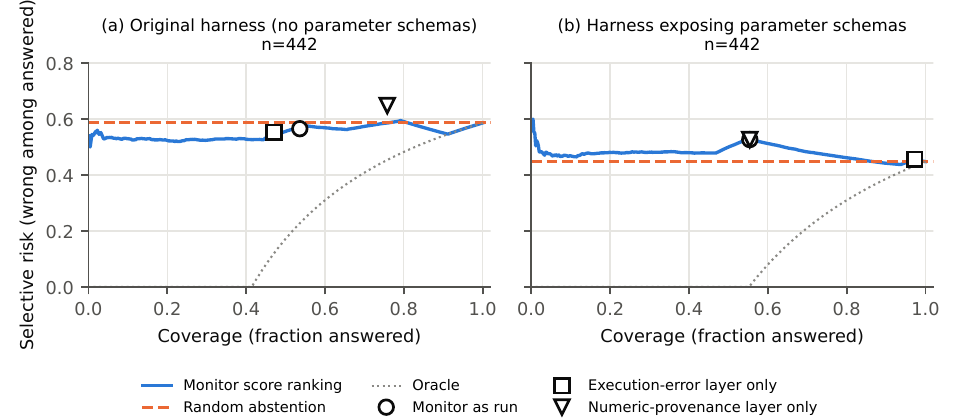}
\caption{Selective risk against coverage for GPT-4o-mini (offline replay, GPT-4o-mini judge labels, 442 scoreable traces). The dashed line is random abstention, the dotted line an oracle. Markers show fixed policies of the monitor.}
\label{fig:riskcoverage}
\end{figure}

Figure~\ref{fig:riskcoverage} shows risk--coverage curves from an offline replay of the monitor on 442 scoreable GPT-4o-mini traces (LLM-judge labels). With the original harness, layer 1 alone stays slightly below random risk (not significant, $p=0.11$); when the harness exposes parameter schemas, malformed calls largely disappear, layer 1 fires on 2.7\% of traces, and the monitor as run is worse than random. The monitor also does not detect the failure studied in RQ2: on 374 schema-valid injections layer 1 fired on 3.2\% of traces against 3.5\% of the same tasks without injection ($p=1.0$), because the corrupted call executes normally. It cannot detect fabrication without a tool call either, since there is no observation to check. Its cost is small: a median of 7.0\,\textmu s per trace (95th percentile 31\,\textmu s) on 4,080 real traces, against a median of 3.6\,s of LLM time per trace.

The monitor is thus a cheap filter for one failure type, execution errors, whose value depends on how often the harness provokes them. It should be evaluated against random abstention, and preferably with a gold-grounded instrument.

\section{Discussion}\label{sec:discussion}

\paragraph{Evaluate the trace, not only the answer.} The main lesson is methodological. When tools are deterministic or can be replayed, the correct observation is known, and correctness can be checked from the trace without a reference text. Reference-based judges answered a different question (is the answer consistent with a description of the task?) and, as a result, ranked models differently, attenuated a large causal effect by a factor of about five to nine, inverted it under the heuristic, and made a monitor look useful. For information systems built around tool-using agents, this suggests logging executed calls and observations and checking answers against them, rather than relying on output-only scoring. The audit of the task files is a second lesson: the stored gold contradicted the query in 38\% of verifiable tasks, so gold specifications themselves need checking against what the agent was asked.

\paragraph{A single wrong argument matters.} When a schema-valid wrong argument reaches execution, two thirds of the tasks that the agent would otherwise have solved end wrong, and agents rarely repeat the call with the correct value. Schema validation does not help, because the argument is valid. Checks that compare executed arguments with the request, for example the city or patient named in the query, would target this failure directly. In the monitor replay such an argument-grounding check separated injected from clean traces, but partly by construction, so we report it only as a direction.

\paragraph{Fabrication is partly an interface problem.} Most fabrication here arises because the harness accepts an action and a final answer in one completion. Native function calling, a stop sequence after the action, or rejecting final answers that follow an unexecuted action would remove much of it. Measuring fabrication remains useful, because a low tool-call rate otherwise looks like robustness in injection experiments.

\paragraph{Changes from versions 1 and 2 of this preprint.}\label{sec:changes}
The re-analysis withdrew several earlier claims: a propagation probability of about 0.62 and the absence of coupling between rejection and recovery (Section~\ref{sec:design}); the statement that the monitor reduces hallucination on every tool-calling model and is net-positive in a precision mode, which reflected abstention volume; a fabrication-classifier precision based on 40 traces; a real-tool pilot of 37 tasks, now 320 pairs; and two-annotator agreement figures, now three annotators with Fleiss' $\kappa$. The open-weight monitor rows, which an intermediate audit could not trace, were recovered from the original run files and reproduce exactly.

\section{Limitations}\label{sec:limitations}

\begin{itemize}
\item \emph{Simulated tools and harness dependence.} Tools are deterministic simulators with small schemas. The harness does not show parameter schemas, which inflates schema errors, and its parser rule inflates fabrication. Tool-call rates, fabrication rates and the size of the paired effect are conditional on this harness; with schemas exposed, GPT-4o-mini's clean-run accuracy under the validator rose from 28.2\% to 54.5\% in a re-run.
\item \emph{Validator development.} The template parsers and normalisation rules were developed while inspecting parts of the same data, not blind. The unit tests, the adjudication of human disagreements and the sensitivity analysis reduce, but do not remove, this risk. Extraction is regex-based and grounding is per task, not per entity.
\item \emph{Human labels.} The 100 human labels were made against descriptive references and cover only one injection condition; the adjudication involved one adjudicator and the gold observations.
\item \emph{Coverage of the causal estimate.} The paired estimate covers seven proprietary models; GPT-4o and Gemini-2.5-Flash have no baseline runs, o3-mini contributes 35 pairs, and each pair is a single run per arm at temperature 0, so run-to-run variation is part of the estimate. Only single-parameter injection into the first call was studied.
\item \emph{Open-weight models.} Their step logs were not kept, so they are scored in answer-only mode, are excluded from the paired analysis, and their monitor results rely on heuristic labels.
\item \emph{Fabrication labels and pilot.} Classifier recall and the judge-based rates are provisional until human labels are collected. The real-tool pilot uses a different task suite and only OpenAI models.
\end{itemize}

\section{Conclusion}\label{sec:conclusion}

For tool-using LLM agents on replayable tools, a deterministic validator grounded in the correct observations changes what the measurements say. Under it, model rankings differ from those of a substring heuristic, a single schema-valid wrong argument raises the error rate by 22.0 points overall and by 44.7 points when the corrupted call executes cleanly, while LLM judges see only a fraction of that effect and a substring heuristic sees the opposite. Fabricated tool use is common in some models and largely produced by the harness. A cheap runtime monitor mostly abstains rather than discriminates. We release the validator, its tests, the traces and the analysis scripts so that these measurements can be repeated and extended to other harnesses and real tools.

\backmatter

\bmhead{Acknowledgements}
\ifanon
\else
The author thanks Yoshitha Challagulla and the third annotator for labelling the calibration traces.
\fi

\section*{Declarations}

\bmhead{Funding} No funding was received for this work.

\bmhead{Competing interests} The author declares no competing interests.

\bmhead{Ethics approval and consent to participate} Not applicable. The study uses synthetic tasks and synthetic patient records; the annotators labelled model outputs and no personal data were collected.

\bmhead{Consent for publication} Not applicable.

\bmhead{Data and code availability}
\ifanon
\else
Tasks, traces, human labels, validator code with unit tests, and all analysis scripts are available at \url{https://github.com/bhaskargurram-ai/agenthallu-bench}. The release accompanying this article adds the open-weight monitor result files and the per-step trace databases, which were not part of the initial public release.
\fi

\bmhead{Use of large language models} LLMs are the objects of study: 13 models were evaluated as agents, and GPT-4o, Gemini-2.5-Flash, GPT-4o-mini, GPT-4.1, GPT-4.1-mini and GPT-5-mini were used as judges whose reliability is analysed. No LLM is part of the structured validator. LLM-based assistants were used for language editing and for help with analysis code; the author checked all code, numbers and text and is responsible for them.

\bmhead{Author contribution}
\ifanon
\else
B.G. is the sole author and carried out the conceptualisation, software, experiments, analysis and writing. ORCID: 0009-0005-2947-102X.
\fi

\bibliography{refs}

\end{document}